%% file: main.tex
\pdfoutput=1
\documentclass{article}

\usepackage{microtype}
\usepackage{graphicx}
\usepackage{booktabs}
\usepackage{amsmath,amssymb}
\usepackage{multirow}
\usepackage{float}
\usepackage[hyphens]{url}
\usepackage{hyperref}

\usepackage{mlsys2025}
\makeatletter
\gdef\isaccepted{1}
\renewcommand{\Notice@String}{Preprint.}
\makeatother
\hypersetup{pdftitle={Is 2:4 Sparsity Still Worth It? Semi-Structured Sparsity versus Quantization in LLM Inference}, pdfsubject={}, pdfauthor={Xiuyi Qin}}

\graphicspath{{figures/}}
\newcommand{\etf}{\ensuremath{e_{2{:}4}}}
\newcommand{\Rpred}{\ensuremath{R_{\mathrm{pred}}}}
\newcommand{\Rbasic}{\ensuremath{R_{\mathrm{basic}}}}
\newcommand{\x}{\ensuremath{\times}}

\mlsystitlerunning{Is 2:4 Sparsity Still Worth It?}

\begin{document}

\twocolumn[
\mlsystitle{Is 2:4 Sparsity Still Worth It? Semi-Structured Sparsity versus Quantization in LLM Inference}

\begin{mlsysauthorlist}
\mlsysauthor{Xiuyi Qin}{aff1}
\end{mlsysauthorlist}

\mlsysaffiliation{aff1}{Independent Researcher, San Jose, CA, USA}
\mlsyscorrespondingauthor{Xiuyi Qin}{xiuyiq@acm.org}

\mlsyskeywords{2:4 sparsity, quantization, sparse tensor cores, LLM inference}

\vskip 0.3in

\begin{abstract}
NVIDIA tensor cores have supported 2:4 semi-structured sparsity since Ampere, and the pattern composes with quantization. When the two are combined, the practical question is which step to take next: from a given precision, add 2:4 sparsity or drop to the next lower precision? We compare the two at equal bits, with sparsity metadata counted, and at equal nominal tensor-core tiers, over all six GEMM operands of a transformer layer, with kernels measured on three GPU generations and online pruning and quantization included. Our results show that as long as a lower precision is available, adding 2:4 is the worse step, both less accurate and slower, because the energy that 2:4 discards from each group of four along the hardware-forced axis does not shrink with bits, while quantization error does. Once quantization reaches the lowest precision the hardware supports, 2:4 still speeds up linear layers by 1.3--1.7\x{}, and on an H100 a retrained 2:4 model served in FP8 by vLLM reaches 27\% higher prefill throughput than the dense FP8 model, with GSM8K accuracy on par. Its quality has to be won back by retraining, however, which restores perplexity long before reasoning; on Blackwell, whose sparse FP4 uses a pairwise 4:8 pattern, the loss is larger and our retraining budget did not recover it.
\end{abstract}
]

\printAffiliationsAndNotice{}

\input{sections/intro}
\input{sections/background}
\input{sections/method}

\input{sections/before}
\input{sections/theory}

\input{sections/floor}
\input{sections/discussion}

\bibliography{refs}
\bibliographystyle{mlsys2025}

\section*{AI Usage Disclosure}
We used Claude (Anthropic), through the Claude Code agent, as a research assistant. It wrote and debugged most of the measurement, kernel and analysis code, ran and monitored the experiments, produced the figures and tables from the result files, and helped condense and revise the text. The author posed the research question and designed the comparison framework, the choice of the next compression step judged at equal bits and at equal tensor-core tiers; decided the measurement protocol, the models, tasks and hardware, and every later change to them; chose which follow-up experiments to run; judged which results support which claims; and wrote the first draft of the paper and determined its structure and argument. The author checked the results and takes full responsibility for the content.

\clearpage
\onecolumn
\appendix
\input{sections/appendix}

\end{document}

%% file: sections/intro.tex
\section{Introduction}
\label{sec:intro}

Semi-structured 2:4 sparsity keeps at most two non-zeros in every group of four consecutive elements along the reduction axis of a matrix multiplication, and NVIDIA's sparse tensor cores run such a multiplication at up to twice the dense rate~\citep{mishra2021}. However, hardware and research on one side and inference software on the other have diverged on whether the feature is worth using. NVIDIA has kept sparse tensor cores in every GPU generation since Ampere and added a sparse FP4 mode in Blackwell, and systems research keeps building on them, for activations~\citep{haziza2025}, for keys and values~\citep{wang2026hierasparse} and for attention scores~\citep{yu2026dynspattn}. In 2026, by contrast, vLLM and LLM Compressor, among the most widely used open-source stacks for compressed LLM inference, removed their 2:4 support because few users relied on it~\citep{vllm36799,llmcompressor011}. A systematic comparison is urgently needed to settle the question.

Because 2:4 sparsity composes with quantization, the choice a practitioner actually faces is about the next step: \textbf{from a given precision, add 2:4 sparsity, or drop to the next lower precision?} The answer depends on whether a lower precision is still available. We call the lowest dense precision that a GPU's tensor cores offer for LLM inference the \emph{quantization floor}: FP8 on Ada and Hopper, NVFP4 on Blackwell. Above the floor, 2:4 has to beat one more step of quantization. At the floor, no such step remains, and this is where the real value of 2:4 emerges.

Existing comparisons of pruning and quantization cannot settle this choice for 2:4. \citet{kuzmin2023} compare unstructured pruning with quantization at nominal compression, mostly for weights, and predict the winner from kurtosis, which measures how heavy-tailed a tensor's values are. \citet{harma2025} show that pruning should precede quantization when the two are combined, and \citet{frantar2025scaling} fold both into a scaling law for pre-training. SLiM~\citep{mozaffari2025slim} reports that 4-bit values with 2:4 sparsity beat 2-bit quantization, but at nominal bits; with its indices counted, that configuration costs 3.1 bits. SLiM also measures its speedup on GPUs, against dense 16-bit models; none of these studies compares adding 2:4 with dropping to a lower dense precision on the hardware. The 2:4 question differs from theirs in four ways. A 2:4 tensor stores a position index next to every kept value, so it costs more than half of its dense counterpart; among prior studies, only \citet{maximov2026} count such metadata, for 8:16 patterns. Sparsity and lower precision each promise twice the tensor-core throughput, but the datasheet says nothing about what each delivers. Most operands at inference time, namely activations, queries, keys, values and attention probabilities, are produced on the fly and must be pruned and compressed before every GEMM, a cost that benchmarks of the sparse GEMM alone leave out. And the hardware fixes which operand of a GEMM can be sparse and along which axis, so statistics of the whole tensor, such as kurtosis, miss what 2:4 actually removes.

We compare the two steps at equal bits, with the 2:4 index counted, and at equal tensor-core tiers (Section~\ref{sec:method}), for each of the six operand families of a transformer layer, sparsified along the axis the hardware forces. The comparison covers every layer of two 8B models and is timed on Ada, Hopper and Blackwell GPUs, including the online pruning and quantization of dynamic operands.

Above the floor, our results show that dropping one precision step is more accurate than adding 2:4 at every storage budget, and that at every tensor-core tier some dense format is both faster and more accurate (Section~\ref{sec:before}). Part of the reason is accounting: 2:4 on $2m$-bit values costs exactly as many bits as dense $(m{+}1)$-bit values, so 2:4 with INT8 values competes with INT5, not INT4. The larger part is the error itself (Section~\ref{sec:theory}). The two values that 2:4 drops from each group of four carry a share of the tensor's energy, the \emph{2:4 energy loss}, that no number of bits for the kept values recovers, whereas quantization error halves with every added bit. Computed from the tensor alone, the 2:4 energy loss tracks the measured gap closely, though it cannot call a close contest, and it explains why Hadamard rotation widens the gap for most operands.

At the floor, 2:4 delivers a real speedup (Section~\ref{sec:floor}). Stacked on FP8, it speeds up linear layers by 1.3--1.7\x{}, and a retrained 2:4 model served by vLLM is faster than the dense FP8 model in prefill and decode, with GSM8K accuracy on par; serving the same weights with dense kernels removes the gain. The price is retraining. One-shot pruning at the floor is unusable, and along our retraining runs perplexity recovers within a quarter billion tokens while GSM8K accuracy stalls far below that of Sparse-Llama, a model retrained thirteen times longer on different data. Perplexity, the common yardstick for retrained sparse models, therefore overstates how much they have recovered. Blackwell's sparse FP4 uses a pairwise 4:8 pattern~\citep{nvidia_cusparselt}: its kernels run 1.4\x{} faster than dense NVFP4, but the pattern discards more of every tensor, cannot reuse models retrained for 2:4, and at our budget does not recover usable reasoning accuracy.

Code, kernels and measurements are available at \url{https://github.com/PuppyQ08/24_quant_compare}.

%% file: sections/background.tex
\section{Background and Related Work}
\label{sec:background}

\paragraph{Sparse tensor cores.} A sparse MMA instruction takes its first operand $A$ in compressed form, the kept values plus a 2-bit index for each, and uses the indices to select the matching elements of the dense operand $B$. The multiplier array thus covers twice the reduction depth per cycle, which bounds the speedup at 2\x{} over the dense instruction of the same precision~\citep{mishra2021}. Microbenchmarks of Hopper measure the instruction itself below that bound~\citep{luo2024hopper}. Three constraints follow, and they shape everything in this paper. Only one operand of a GEMM can be sparse. Its groups of four must lie along the reduction axis. And the speedup applies to the multiplication, not to whatever produces the compressed operand. For weights, the standard recipe prunes a trained model and retrains it with the mask fixed~\citep{mishra2021,pool2021,zhou2021nm}; for LLMs, SparseGPT~\citep{frantar2023sparsegpt} and Wanda~\citep{sun2024wanda} prune in one shot, and Sparse-Llama retrains Llama-3.1-8B to 2:4 on 13B tokens~\citep{marques2024sparsellama}. For FP4 data, Blackwell's sparse tensor cores use a paired 4:8 pattern instead of 2:4~\citep{nvidia_cusparselt}.

\paragraph{Quantization.} Integer formats store $k$-bit values with a shared scale per group of elements, typically 128. FP8 (E4M3) needs only one scale per row or channel. Block-scaled 4-bit floating-point formats store E2M1 values with a shared scale per small block: MXFP4 uses a power-of-two scale per 32 elements~\citep{rouhani2023mx}, and NVFP4 an E4M3 scale per 16 elements plus a per-tensor scale. The tensor cores of Ada and Hopper go down to 8-bit floating point; Blackwell adds FP4 at twice the FP8 rate. On top of these formats, weights are quantized with GPTQ~\citep{frantar2023gptq} and AWQ~\citep{lin2024awq}, activations together with weights with LLM.int8()~\citep{dettmers2022int8} and SmoothQuant~\citep{xiao2023smoothquant}, and KV caches with KIVI~\citep{liu2024kivi} and KVQuant~\citep{hooper2024kvquant}; attention runs in low precision in FlashAttention-3~\citep{shah2024fa3} and SageAttention~\citep{zhang2025sage}, and rotations such as QuaRot~\citep{ashkboos2024quarot} and SpinQuant~\citep{liu2025spinquant} remove outliers before low-bit quantization.

\paragraph{Sparsity beyond weights.} Activation sparsity is exploited through ReLU-like activations~\citep{mirzadeh2024relu,song2025prosparse}, training-free thresholds~\citep{liu2025teal} and N:M patterns~\citep{haziza2025,an2025amber,alanova2026}. For attention, HieraSparse~\citep{wang2026hierasparse} applies 2:4 to keys and values, DynSpAttn~\citep{yu2026dynspattn} and DFSS~\citep{chen2023dfss} to scores, and MUSTAFAR~\citep{joo2025mustafar} prunes the KV cache without structure. Each of these systems picks one operand and one axis; our measurements put all of them on one scale and against the quantization each would otherwise use.

\paragraph{The quantization floor.} Besides FP8 and FP4, Ada also has INT4 tensor cores, which LLM inference stacks do not use, and we leave them out; the floor is thus a property of the formats inference stacks run, not of the silicon alone. For storage alone, we take 4-bit weights as the practical floor.

%% file: sections/method.tex
\section{Methodology: Comparing on Equal Terms}
\label{sec:method}

Adding 2:4 sparsity and dropping to a lower precision both serve one of two purposes: saving memory or saving compute. We compare the two at equal cost for each purpose. At \emph{equal bits} they occupy the same memory, so the only question is which is more accurate. At \emph{equal tensor-core tiers} they promise the same speedup, so we ask both which is more accurate and which is actually faster.

\subsection{Equal bits: the same memory}
\label{sec:bits}

\begin{figure}[t]
\centering
\includegraphics[width=\columnwidth]{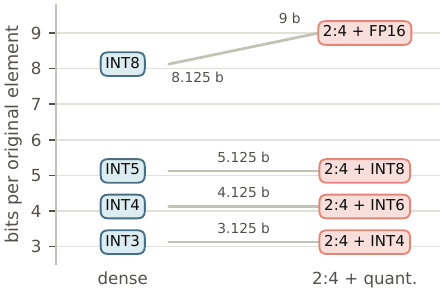}
\caption{\textbf{Equal bits.} Bits per original element, including the 2:4 index (1 bit) and one FP16 scale per 128 elements. Each line joins two formats of identical cost; 2:4 on FP16 values costs more than FP8.}
\label{fig:bits}
\end{figure}

We charge each format for every bit it stores per original element: values, indices, scales and zero points. Dense symmetric INT$k$ with one FP16 scale per $g$ elements costs $k + 16/g$ bits. A 2:4 tensor keeps half of the elements, so with INT$k'$ values it stores $k'/2$ value bits and one index bit per original element, plus the same scales, one per $g$ original elements. Writing $k' = 2m$ gives
\begin{equation}
\text{bits}\big(\text{2:4} + \text{INT}(2m)\big) \;=\; m + 1 + \tfrac{16}{g} \;=\; \text{bits}\big(\text{INT}(m{+}1)\big).
\label{eq:identity}
\end{equation}
Dropping half of the elements thus saves far less than half of the bits. 2:4 with INT8 values costs as much as dense INT5, 2:4 with INT4 values as much as INT3, and 2:4 with FP16 values costs more than FP8 (Figure~\ref{fig:bits}). At equal bits, the real question is whether to drop half of the elements and store the rest with twice the precision, or to keep all of them with one more bit each.

Storage budgets apply to the operands that live in memory: weights, keys and values. Storage is not tied to tensor-core precisions, since weights and caches can be dequantized before the GEMM, and integer formats with group scales are the standard choice for both~\citep{frantar2023gptq,lin2024awq,liu2024kivi,hooper2024kvquant}. We therefore state the identity for integers and compare at four budgets from 9 down to 3.1 bits per element; floating-point formats join wherever their cost fits and are the strongest rivals at the higher budgets. Each 2:4 configuration meets only dense formats that cost the same or less, except MXFP4, an eighth of a bit more expensive than its 2:4 rival (Appendix~\ref{app:bits}).

\subsection{Equal tiers: the same promised speed}
\label{sec:tiers}

For compute, we pair formats that promise the same \emph{tensor-core tier}, the nominal throughput relative to dense 16-bit GEMMs given on the datasheet~\citep{nvidia_l40s,nvidia_h100}. Halving precision doubles the nominal throughput, and so does 2:4 sparsity. In the \emph{2\x{} tier}, a GEMM with one 2:4 operand in FP16 (BF16 runs at the same rate) meets a dense GEMM with both operands in FP8 or INT8. In the \emph{4\x{} tier}, 2:4 with FP8 operands meets dense 4-bit floating point, which only Blackwell runs; on Hopper, 2:4 is the only way into this tier. The \emph{8\x{} tier}, sparse NVFP4 against dense NVFP4, lies at the floor and belongs to Section~\ref{sec:floor}.

In the 2\x{} tier the dense side quantizes both operands of the GEMM, while 2:4 alters only one. A tier states only the nominal throughput, so we also measure what each format actually delivers (Section~\ref{sec:speed}).

\subsection{Operands and forced axes}
\label{sec:operands}

\begin{figure*}[t]
\centering
\includegraphics[width=\textwidth]{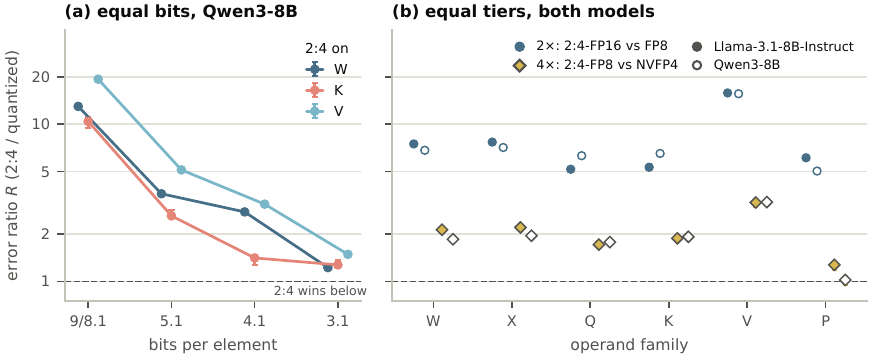}
\caption{\textbf{Accuracy above the floor.} Layer-median error ratio $R$ (2:4 over quantized; below 1, 2:4 would win), 95\% bootstrap intervals. (a) Equal bits on Qwen3-8B against the strongest quantizer of each budget (Llama: Figure~\ref{fig:bits_llama}). (b) Equal tiers on both models. All values: Tables~\ref{tab:storage_all} and~\ref{tab:compute_all}.}
\label{fig:accuracy}
\end{figure*}

\begin{table}[t]
\centering
\caption{\textbf{The six operand families}, the GEMM in which each is the sparse operand, and the axis along which the hardware forces its 2:4 groups. Only weights are static; all other operands are produced at inference time.}
\label{tab:operands}
\resizebox{\columnwidth}{!}{%
\begin{tabular}{lllll}
\toprule
& Tensor & GEMM & Forced axis & Static \\
\midrule
W & linear weights & $Y = XW^\top$ & input features & yes \\
X & linear inputs & $Y = XW^\top$ & hidden & no \\
Q & queries & $S = QK^\top$ & head dim & no \\
K & keys & $S^\top = KQ^\top$ & head dim & no \\
V & values & $O^\top = V^\top P^\top$ & tokens & no \\
P & attention probs. & $O = PV$ & keys & no \\
\bottomrule
\end{tabular}}
\end{table}

A transformer layer has six families of GEMM operands that could be sparse (Table~\ref{tab:operands}). Only one operand per GEMM can be sparse, and that GEMM's reduction axis fixes the axis of its 2:4 groups. Weights are grouped along input features, queries and keys along the head dimension, values along tokens and attention probabilities along keys. Only weights can be pruned offline and retrained. For attention probabilities we follow DynSpAttn~\citep{yu2026dynspattn} and keep the two largest of every four scores before the softmax, which then normalizes over the kept entries.

\subsection{Measuring accuracy}
\label{sec:R}

The primary metric is operator-level relative error against an FP32 reference: the output of a linear layer with real calibration activations, or the output of an attention head. We measure it for every \emph{cell}, a combination of model, operand family, budget or tier, and rival format, and define
\begin{equation}
R \;=\; \frac{\text{error of the best 2:4 configuration}}{\text{error of the best quantized configuration}},
\label{eq:R}
\end{equation}
so that $R < 1$ means 2:4 wins. Each side picks from an equal-effort menu. The 2:4 side tries magnitude pruning, Wanda and SparseGPT for weights, and per-token shift and variance correction~\citep{alanova2026} for dynamic operands. The quantized side tries round-to-nearest along its best grouping axis, GPTQ~\citep{frantar2023gptq} and AWQ~\citep{lin2024awq}. Combined formats prune first and then quantize the kept values with fresh scales~\citep{harma2025}. Both sides calibrate on the same C4 sequences~\citep{raffel2020c4}.

We aggregate in the log domain, so that ``twice as bad'' and ``twice as good'' weigh the same: the layer-median $R$ of a cell is the median over layers of the per-layer median over evaluation samples, and its confidence interval comes from a paired bootstrap over samples (Appendix~\ref{app:atlas}). Operator error ignores how errors compound across layers, so we also check selected cells at model level, compressing the family in every layer and measuring WikiText-2 perplexity~\citep{merity2017wikitext}, GSM8K accuracy~\citep{cobbe2021gsm8k} and, for keys, values and probabilities, RULER at 16K tokens~\citep{hsieh2024ruler}.

\subsection{Measuring speed}
\label{sec:timing}

\begin{figure*}[t]
\centering
\includegraphics[width=\textwidth]{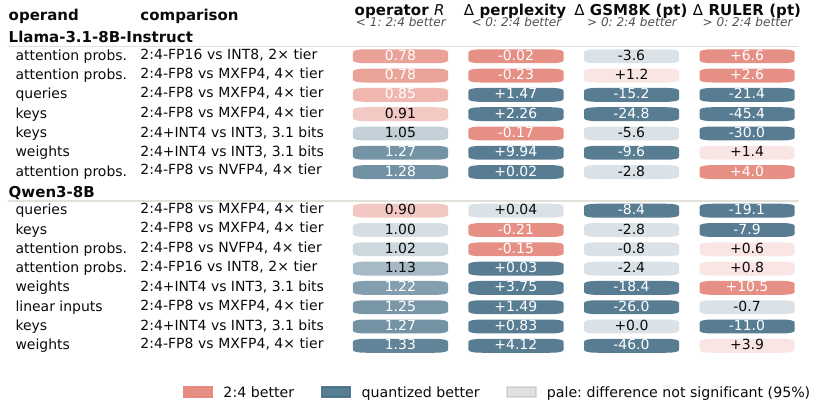}
\caption{\textbf{Model quality on the cells closest to parity} ($0.7 < R < 1.4$), sorted by operator-level $R$; each cell compresses one operand family in every layer. Changes are 2:4 minus quantized. Coral: 2:4 better; blue: quantized better; pale: not significant at 95\%. 2\x{}, 4\x{}: tensor-core tiers; 3.1 bits: storage budget.}
\label{fig:model}
\end{figure*}

We time on AWS cloud instances with an L40S (Ada), an H100 (Hopper) and an RTX PRO 6000 (Blackwell, sm\_120), using device kernel time from CUPTI on the linear-layer shapes of both models at prefill lengths. Sparse GEMMs use cuSPARSELt with an exhaustive search over algorithm and split-K; dense GEMMs use the library's default heuristics.

Each side pays for what it needs at run time. Weights are static: pruning, compression and quantization happen offline, and only the GEMM is timed. Every other operand is dynamic, and each side pays for producing its own input. The 2:4 side must select two of every four elements and write the kept values and their indices, optionally quantized; the dense side must quantize. For the 2:4 side we take the faster of cuSPARSELt's own prune-and-compress and a fused kernel that reads the operand once and writes the compressed form once. The fused kernel reimplements the HieraSparse kernel~\citep{wang2026hierasparse} in Triton, bit for bit, and optionally fuses FP8 or INT8 quantization. The dense side uses fused per-token FP8 and INT8 quantization and an NVFP4 block quantizer. Every kernel matches a PyTorch reference bit for bit. The per-token shift and variance correction, which win on accuracy for dynamic operands, are not included in the 2:4 timing.

Operands are resident in L2, as in a pipeline where the previous kernel has just written them; preprocessing is also timed with L2 flushed, with the same conclusions (Appendix~\ref{app:timing}). Whole models are served in vLLM for an end-to-end check (Section~\ref{sec:speed2}).

%% file: sections/before.tex
\section{Above the Floor, One More Step of Quantization Wins}
\label{sec:before}

We now run the comparison over all of our measurements: Llama-3.1-8B-Instruct~\citep{grattafiori2024llama3} and Qwen3-8B~\citep{yang2025qwen3}, every layer, all six operand families, four storage budgets and two compute tiers, over 220{,}000 measured configurations in total.

\subsection{Accuracy}
\label{sec:accuracy}

\begin{figure}[t]
\centering
\includegraphics[width=\columnwidth]{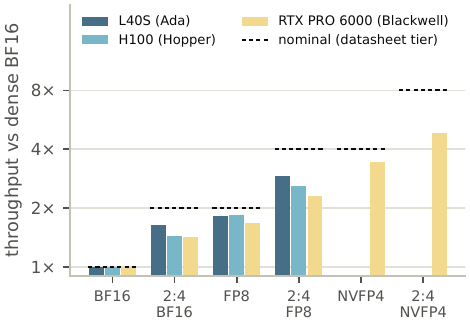}
\caption{\textbf{Nominal against measured throughput} of linear-layer GEMMs relative to dense BF16 (sparse weights, kernel time, median over prefill shapes). Dashed: the nominal tier. NVFP4 and its sparse variant exist only on Blackwell.}
\label{fig:throughput}
\end{figure}

\begin{table}[t]
\centering
\caption{\textbf{Speed at equal tier and at the floor.} $t(\text{rival}) / t(\text{2:4 side})$, above 1 means 2:4 is faster; median over the Llama-3.1-8B projections at 2K--16K tokens (ranges: Table~\ref{tab:hw_tiers}). W: sparse weights, kernel time; X: sparse activations, preprocessing on both sides. Floor: 2:4 over the dense floor format.}
\label{tab:speed}
\resizebox{\columnwidth}{!}{\input{tables/speed}}
\end{table}

\paragraph{Equal bits.} 2:4 loses at every storage budget on both models (Figure~\ref{fig:accuracy}a for Qwen3, Figure~\ref{fig:bits_llama} for Llama). The gap is more than an order of magnitude at 8 bits and shrinks steadily as bits fall, but it does not close: at 3.1 bits the closest cell, Llama's keys against INT3, is still just above parity.

\begin{figure*}[t]
\centering
\includegraphics[width=\textwidth]{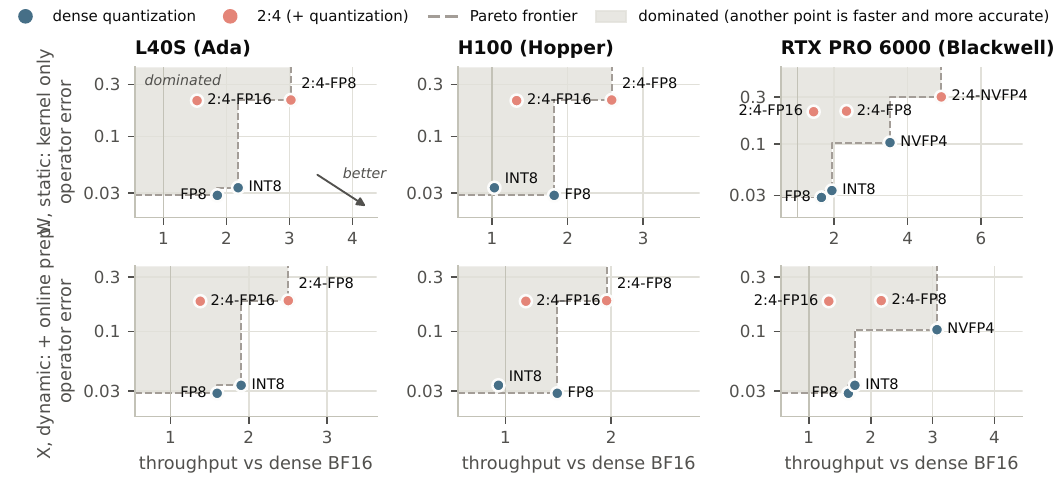}
\caption{\textbf{Error against throughput for linear layers} (Qwen3-8B; Llama: Figure~\ref{fig:pareto_llama}). Each point is one format: throughput relative to dense BF16 (prefill median) against layer-median operator error; lower right is better. Gray: dominated (another format is faster and more accurate); dashed: Pareto frontier. Top: sparse weights; bottom: sparse activations, online preprocessing on both sides.}
\label{fig:pareto}
\end{figure*}

\paragraph{Equal tiers.} In the 2\x{} tier, 2:4 with FP16 values has five to sixteen times the error of dense FP8, for every operand family on both models (Figure~\ref{fig:accuracy}b). In the 4\x{} tier, 2:4 with FP8 values has at least the error of dense NVFP4 for every family, and attention probabilities come closest. Of the 108 compute cells, 2:4 wins six, all against weak rivals: MXFP4, whose power-of-two block scales waste precision, and row-wise INT8 for attention probabilities. Against FP8 and NVFP4 it wins none.

\paragraph{Model quality.} The hardest test for our conclusion at model level uses the fifteen cells whose operator-level ratio is closest to parity, the cells most favourable to 2:4 (Figure~\ref{fig:model}). On GSM8K, 2:4 is never significantly better than quantization, and it is significantly worse in seven of these cells, including three of the five in which it wins at operator level. 2:4 does significantly better on perplexity or RULER in nine cases. Six are attention probabilities, where keeping the top two of every four scores sharpens attention: it helps long-context retrieval and perplexity, but never reasoning. The other three trade one metric for another: small perplexity gains on keys that RULER contradicts, and better retrieval for weights against INT3 on Qwen3 at the cost of eighteen GSM8K points. Near parity, the operator-level winner is a weak guide, matched by the three metrics in just over half of the cases; far from parity, it matches the perplexity outcome on eleven of twelve representative cells (Table~\ref{tab:proxy}).

\paragraph{Why operator error and model quality can disagree.} Three differences between a per-layer ratio and a whole model explain the disagreements. First, $R$ is a median over layers, but a few layers can decide model quality. In Llama's queries and keys, 2:4 beats MXFP4 in about three quarters of the layers, yet has two to three times its error in the first two layers (Figure~\ref{fig:layers}), and whatever goes wrong there passes through every later layer. Second, operator error is measured with clean inputs, while the model compresses every layer at once, so errors compound through the residual stream and the softmax; Qwen3 has no such outlier layers, and its query cell still reverses at model level. Third, the form of an error matters, not only its size. 2:4 sets whole components to zero, whereas quantization spreads small rounding errors over all of them, and the tasks react differently: retrieval in RULER depends on sharp attention peaks, which pruning queries and keys disturbs and pruning probabilities sharpens, while GSM8K accumulates every small error over long chains of generated tokens. Near parity, these effects outweigh a small difference in operator error; far from parity, they do not.

\subsection{Speed}
\label{sec:speed}

Figure~\ref{fig:throughput} shows what each tier delivers: on all three GPUs, dense FP8 realizes more of its nominal 2\x{} than 2:4 with 16-bit values does. Table~\ref{tab:speed} confirms it at equal tier. With sparse weights, 2:4 with FP16 values runs slower than dense FP8 on all three GPUs, and on Blackwell 2:4 with FP8 values runs slower than dense NVFP4. With sparse activations, both sides pay for preprocessing and the ratios barely move; preprocessing matters more against the same precision, where only 2:4 pays it (Tables~\ref{tab:hw_tiers} and~\ref{tab:prepshare}). The one exception, FP16 against INT8 on H100, reflects a weak INT8 GEMM on that GPU, where INT8's error is a ninth of 2:4's. The ordering carries over to a serving engine: in vLLM on an H100, dense FP8 serves faster than 2:4 with BF16 values, in prefill and in decode, from a smaller model (Table~\ref{tab:e2e}).

How a dynamic operand is pruned matters as much as the sparse GEMM. A fused kernel that reads the operand once and writes the kept values and indices once runs close to memory copy bandwidth, whereas cuSPARSELt's own prune-and-compress path and PyTorch's semi-structured conversion are several times slower (Table~\ref{tab:prep}) and give away most of what the sparse GEMM gains. Attention shows the limits of 2:4 on dynamic operands (Table~\ref{tab:attention}): queries and keys, reduced over a 128-channel head, roughly break even at equal precision, and for attention probabilities, pruning and compressing $P$ online leaves 2:4 at about half the dense speed, which rules out $P$ as a standalone sparse operand.

\subsection{Accuracy and speed together}
\label{sec:pareto}

Figure~\ref{fig:pareto} puts the two together. Above the floor, every 2:4 configuration falls in the dominated region: some dense format is both faster and several times more accurate. 2:4 reaches the Pareto frontier only at its fast end, after it has been stacked on the floor, which is the subject of Section~\ref{sec:floor}.

%% file: tables/speed.tex
\begin{tabular}{lcccccc}
\toprule
 & \multicolumn{2}{c}{2$\times$: vs FP8} & \multicolumn{2}{c}{4$\times$: vs NVFP4} & \multicolumn{2}{c}{floor: 2:4 added} \\
\cmidrule(lr){2-3}\cmidrule(lr){4-5}\cmidrule(lr){6-7}
GPU & W & X & W & X & FP8 & NVFP4 \\
\midrule
L40S & 0.89 & 0.89 & -- & -- & 1.65 & -- \\
H100 & 0.74 & 0.80 & -- & -- & 1.42 & -- \\
RTX PRO 6000 & 0.84 & 0.85 & 0.66 & 0.70 & 1.33 & 1.40 \\
\bottomrule
\end{tabular}

%% file: sections/theory.tex
\section{Analysis: Why 2:4 Loses}
\label{sec:theory}

\begin{figure*}[t]
\centering
\includegraphics[width=\textwidth]{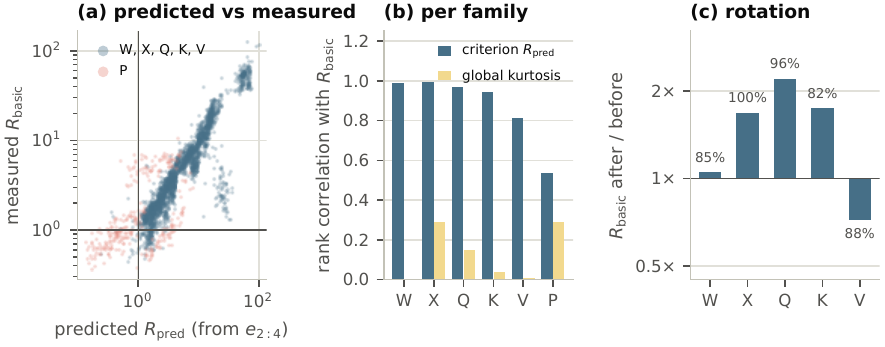}
\caption{\textbf{The criterion.} (a) Predicted \Rpred{} against measured \Rbasic{} for every (model, layer, cell) point. (b) Rank correlation with \Rbasic{} per operand family, for the criterion and for global kurtosis. (c) Change of \Rbasic{} under Hadamard rotation (block 128, both operands), median per family; numbers: share of points whose direction the criterion predicts.}
\label{fig:criterion}
\end{figure*}

\begin{table}[t]
\centering
\caption{\textbf{The asymmetry on Gaussian data.} Relative error of magnitude 2:4 (kept values quantized with round-to-nearest) and of the dense format of equal bits, for i.i.d.\ Gaussian data with one scale per 128 elements.}
\label{tab:gauss}
\small
\input{tables/gauss}
\end{table}

The results of Section~\ref{sec:before} follow from a simple asymmetry. Quantization keeps every value and rounds it, and its error shrinks quickly as bits are added. 2:4 sparsity deletes half of the values; the deleted values are lost entirely, and no number of bits spent on the kept ones brings them back. Table~\ref{tab:gauss} shows the effect on Gaussian data at equal bits: the error of 2:4 barely moves across budgets, while the error of the dense rival doubles with every bit removed, so the two would meet only below 3 bits, a precision no dense tensor core supports. The rest of this section makes the argument precise for any tensor.

\paragraph{The energy that 2:4 removes.} Let $A$ be the operand that 2:4 prunes, the sparse operand of its GEMM in Table~\ref{tab:operands}, such as the weight matrix of a linear layer or the key matrix of an attention head. Magnitude 2:4 keeps the two largest of every four elements of $A$ along the forced axis and drops the other two. Its relative error energy is the share of energy in the dropped elements, which we call the \emph{2:4 energy loss}:
\begin{equation}
\etf(A) \;=\; \frac{\sum_{g} \big(a_{g1}^2 + a_{g2}^2\big)}{\|A\|_F^2},
\label{eq:rho4}
\end{equation}
where $g$ runs over the groups of four along the forced axis, $a_{g1}$ and $a_{g2}$ are the two entries of smallest magnitude in group $g$, and $\|A\|_F^2$, the sum of all squared entries of $A$, is its total energy. For i.i.d.\ Gaussian data $\etf = 0.132$. It is small when each group is dominated by one or two large entries and large when the four entries are alike. Unlike kurtosis, \etf{} depends on the axis: the same tensor can have very different values along its rows and its columns, and only the forced axis matters to the hardware.

\paragraph{Equal bits.} Once the mask is fixed, pruning and quantization act on disjoint elements: the dropped elements carry the pruning error, and the kept ones carry the quantization error. Their error energies therefore add exactly. In the same notation, let $e_k(A)$ be the relative error energy of a $k$-bit quantizer on $A$, and $e_{k}^{\text{kept}}(A)$ the error of quantizing only the kept elements, normalized by the energy of all of $A$. At the equal-bit pairing of Equation~\eqref{eq:identity}, the ratio of tensor errors is
\begin{equation}
\Rpred^2 \;=\; \frac{\etf(A) + e_{2m}^{\text{kept}}(A)}{e_{m+1}(A)},
\label{eq:rrule_bits}
\end{equation}
and 2:4 wins exactly when
\begin{equation}
\etf(A) \;<\; e_{m+1}(A) - e_{2m}^{\text{kept}}(A).
\label{eq:win}
\end{equation}
The right-hand side shrinks by roughly a factor of four with every bit added to the dense format, because each bit halves the quantization step. The left-hand side does not depend on bits at all. 2:4 is therefore far behind at 8 bits, closer at 4, and can only win once dense quantization error grows to the size of the pruning loss, which for Gaussian data happens only below 3 bits, beneath anything dense tensor cores offer. The inequality also says where help for 2:4 has to come from: not from a better quantizer for the kept values, but from a tensor whose dropped entries carry little energy, or a model trained to do without them, which is what retraining provides at the floor (Section~\ref{sec:floor}).

The same reasoning gives the predicted ratio at the compute tiers, where both operands of the GEMM contribute error (Appendix~\ref{app:tiers}). We call \Rpred{}, the predicted error ratio, the criterion. Equation~\eqref{eq:rrule_bits} is exact for the tensors and contains no fitted parameters: \etf{} and the quantization errors are computed from the operands directly. For the GEMM output it is an approximation, because the output error also depends on the other operand. For $Y = XW^\top$, a weight error $E$ gives the output error energy
\begin{equation}
\mathbb{E}\,\|XE^\top\|_F^2 \;=\; \operatorname{tr}\!\big(E\,\Sigma_X E^\top\big),
\label{eq:weighted}
\end{equation}
where $\Sigma_X$ is the second-moment matrix of the inputs, which equals the tensor error only for isotropic inputs. The criterion ignores this weighting; Section~\ref{sec:criterion} tests how much that matters.

\paragraph{Consequences.} The analysis has three testable consequences. First, the gap narrows as bits fall but does not close at the precisions tensor cores support. Second, which operands come closest depends on their structure along the forced axis, not on their overall distribution, so global kurtosis should predict the outcome poorly. Third, Hadamard rotation~\citep{ashkboos2024quarot,liu2025spinquant} changes both sides: it shrinks the quantizer's range and makes the four entries of each group more alike, which raises \etf{}. Attention probabilities lie outside the model, because most of their energy sits in a few large probabilities that 2:4 keeps, while the output loses the many small ones it drops.

\subsection{Testing the analysis}
\label{sec:criterion}

\begin{figure*}[t]
\centering
\includegraphics[width=\textwidth]{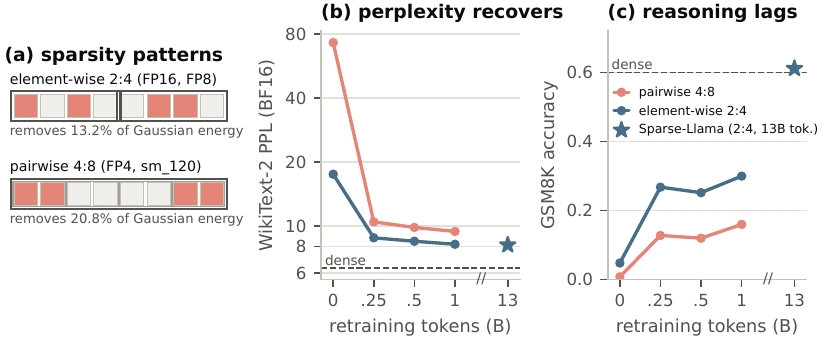}
\caption{\textbf{2:4 at the floor} (Llama-3.1-8B). (a) Kept elements (orange) along eight elements of the reduction axis: element-wise 2:4, and the pairwise 4:8 of Blackwell's sparse FP4. (b) BF16 WikiText-2 perplexity and (c) GSM8K accuracy along the retraining of our two students, from the one-shot SparseGPT start (0), with Sparse-Llama, retrained for 13B tokens, for reference. Perplexity keeps falling while GSM8K stalls.}
\label{fig:floor}
\end{figure*}

\begin{table*}[t]
\centering
\caption{\textbf{The cost of stacking 2:4 on a quantization floor} (Llama-3.1-8B; WikiText-2 perplexity / GSM8K accuracy on 250 problems). Retrained 2:4 (13B tokens) is Sparse-Llama~\citep{marques2024sparsellama}, quantized with its mask fixed. Ours (1B tokens): element-wise 2:4 except in the hardware NVFP4 row, which uses pairwise 4:8 (4:8p). $^\dagger$Element-wise sparse NVFP4 does not exist in hardware and is shown for reference.}
\label{tab:stacking}
\resizebox{\textwidth}{!}{\input{tables/stacking}}
\end{table*}

Equations~\eqref{eq:rrule_bits} and~\eqref{eq:win} describe the two basic methods, magnitude pruning and round-to-nearest. We therefore test them against \Rbasic{}, the error ratio measured at the GEMM output when each side uses its basic method; calibrated methods such as SparseGPT and GPTQ compensate errors in ways that statistics of the tensor do not capture.

The first consequence holds: the measured storage ratios of Figure~\ref{fig:accuracy} fall with bits and stay above parity, as the Gaussian ones do; the heavier tails of real tensors help 2:4 at high budgets, but nowhere near enough.

Point by point, the criterion tracks the size of the measured gap closely (Figure~\ref{fig:criterion}a). For weights, activations, queries, keys and values, it ranks the points with a correlation of 0.93 and, with no parameters to fit, separates the points where 2:4 wins with an AUC of 0.95. It is not a classifier at the threshold, however (Table~\ref{tab:criterion}): 2:4 wins at only 3\% of these points, so always predicting quantization is as accurate, and the criterion recognizes only one in eight of 2:4's wins, because its prediction runs high near parity, most visibly for queries and keys against MXFP4. It explains how large the gap is and why, not who wins a close call, which is settled at model level (Section~\ref{sec:accuracy}). Global kurtosis, the predictor proposed by \citet{kuzmin2023}, barely correlates with the outcome (Figure~\ref{fig:criterion}b), and fitted on one model, it separates winners from losers on the other far worse than the parameter-free criterion does. What decides the comparison is the structure of a tensor inside the groups the hardware prunes, not its overall shape.

Rotation tilts the comparison further toward quantization (Figure~\ref{fig:criterion}c). Hadamard rotation roughly doubles the error ratio for queries, keys and linear-layer inputs, leaves weights almost unchanged, and favours 2:4 only for values. The criterion predicts the direction of change at 87\% of the points, against 68\% for always predicting an increase. Since rotation is now a standard step before low-bit quantization, the deployed comparison is less favourable to 2:4 than our unrotated measurements.

The criterion does not hold for attention probabilities, as the analysis anticipates; at model level, 2:4 on probabilities even moves perplexity in opposite directions on the two models. We treat probabilities as a form of attention sparsification rather than compression.

%% file: tables/gauss.tex
\begin{tabular}{lllll}
\toprule
Bits & 2:4 side & error & dense rival & error \\
\midrule
9 / 8.1 & 2:4 + FP16 & 36.4\% & INT8 & 0.6\% \\
5.1 & 2:4 + INT8 & 36.4\% & INT5 & 5.5\% \\
4.1 & 2:4 + INT6 & 36.4\% & INT4 & 11.7\% \\
3.1 & 2:4 + INT4 & 37.3\% & INT3 & 27.4\% \\
\bottomrule
\end{tabular}

%% file: tables/stacking.tex
\begin{tabular}{lllll}
\toprule
Floor & Dense & + one-shot 2:4 & + retrained 2:4 (13B tok.) & + retrained, ours (1B tok.) \\
\midrule
BF16 & 6.31 / 0.600 & -- & 8.13 / 0.612 & 8.19 / 0.300 (2:4), 9.41 / 0.160 (4:8p) \\
INT4 weights (W4A16) & 6.90 / 0.540 & 18.61 / 0.032 & 8.45 / 0.580 & 8.53 / 0.292 \\
FP8 (W8A8) & 6.36 / 0.552 & 17.82 / 0.028 & 8.19 / 0.632 & 8.25 / 0.308 \\
NVFP4 (W4A4), pairwise 4:8 & 6.98 / 0.460 & 108.83 / 0.020 & -- & 10.22 / 0.132 \\
NVFP4, element-wise 2:4$^\dagger$ & 6.98 / 0.460 & 21.74 / 0.020 & 8.72 / 0.516 & 8.82 / 0.192 \\
\bottomrule
\end{tabular}

%% file: sections/floor.tex
\section{2:4 at the Quantization Floor}
\label{sec:floor}

At the floor there is no lower dense format to switch to, and 2:4 can only be added on top. We study three floors for the linear layers, with attention kept in BF16: FP8 weights and activations, the floor of Ada and Hopper; NVFP4 weights and activations, the floor of Blackwell; and 4-bit integer weights, the practical storage floor. 2:4 is applied to the weights, which can be pruned offline and retrained. This section uses the base Llama-3.1-8B, from which both Sparse-Llama and our retrained students are derived.

\subsection{How much faster}
\label{sec:speed2}

Stacked on FP8, 2:4 speeds up prefill-shaped linear layers by 1.3 to 1.7\x{}, depending on the GPU (Table~\ref{tab:speed}), which lifts their throughput to between two and three times dense BF16. On Blackwell, sparse NVFP4 runs 1.4\x{} faster than dense NVFP4, for nearly five times dense BF16 (Figure~\ref{fig:throughput}). These gains belong to the compute-bound regime. At decode, GEMMs are bound by memory bandwidth and 2:4 saves only bytes, which helps on some GPUs and hurts on others (Table~\ref{tab:decode} and Figure~\ref{fig:roofline}). We did not time the 4-bit weight floor, which serves decode and whose sparse kernels vLLM has since removed~\citep{vllm32688}.

\begin{table}[t]
\centering
\caption{\textbf{End-to-end serving} in vLLM on one H100 (dense: Llama-3.1-8B; 2:4: Sparse-Llama). Prefill: 64 prompts of 2048 tokens; decode: 256 tokens at batch $b$. No GSM8K difference is significant.}
\label{tab:e2e}
\small
\setlength{\tabcolsep}{4pt}
\input{tables/e2e}
\end{table}

The gain survives in a serving engine (Table~\ref{tab:e2e}). Served with vLLM~\citep{kwon2023vllm} from LLM Compressor checkpoints (Appendix~\ref{app:timing}), 2:4 on the FP8 floor raises prefill and decode throughput, at batch 1 as at batch 64, and shrinks the weights in memory by more than a quarter. Serving the same weights with dense kernels removes the speedup and leaves GSM8K unchanged, so the gain comes from the sparse kernels alone; it is smaller than at the GEMM level because attention and the other layers do not shrink. With BF16 values, the engine's 2:4 path is no faster than dense BF16, well short of cuSPARSELt on the same GPU.

\subsection{What pairwise 4:8 costs on Blackwell}
\label{sec:pair48}

For FP4 data, Blackwell's sparse tensor cores do not use element-wise 2:4. cuSPARSELt documents a paired 4:8 pattern~\citep{nvidia_cusparselt}: in each group of eight elements, two of the four adjacent pairs are kept (Figure~\ref{fig:floor}a), and in the block-scaled GEMM both operands carry one scale per 32 positions of the reduction axis, half the scale resolution of dense NVFP4. We checked the format on an RTX PRO 6000 (sm\_120) with cuSPARSELt 0.8.1: element-wise masks give wrong products, while the pairwise format matches our simulation to within rounding; sparse FP8 on the same GPU remains element-wise. What the documentation does not say is what the pattern costs in an LLM.

Pairing costs quality. Keeping pairs rather than elements removes about one and a half times as much energy from Gaussian data, and one-shot pruning to pairs at the NVFP4 floor gives Llama-3.1-8B five times the perplexity that element-wise pruning would (Table~\ref{tab:stacking}). Almost all of this comes from the pattern, not from the coarser scales, which barely matter under either mask (Table~\ref{tab:fp4ablation}). Nor can a model retrained for element-wise 2:4 be reused: converting Sparse-Llama to pairs discards a third of its trained weights and destroys it.

\subsection{What it costs, and what retraining buys}
\label{sec:cost}

Table~\ref{tab:stacking} gives the cost of stacking 2:4 on each floor. One-shot pruning is unusable on every floor: perplexity more than doubles and GSM8K accuracy collapses to a few percent. Instruction-tuned Llama and Qwen3 behave alike.

Retraining recovers most of the loss. With Sparse-Llama, the general 2:4 model retrained for 13B tokens~\citep{kurtic2025sparsellama}, the perplexity increase on the INT4 and FP8 floors shrinks to a small fraction of the one-shot increase, and GSM8K accuracy fully returns. No such model exists for the pairwise format, so we retrained two students, one with pairwise 4:8 masks and one with element-wise 2:4 masks as a control, by distillation from the dense model with the one-shot SparseGPT masks fixed~\citep{kurtic2023squarehead}, on the same 1B tokens of FineWeb-Edu~\citep{penedo2024fineweb} with one untuned recipe (Table~\ref{tab:recipe}).

Two results stand out (Figure~\ref{fig:floor}b,c). First, the pairwise format costs more than element-wise 2:4 throughout training, and the gap is still clear at 1B tokens. Second, perplexity overstates what retraining has achieved. Along both runs, perplexity keeps falling, while GSM8K accuracy jumps within the first quarter billion tokens and then barely moves. At 1B tokens, the pairwise student has recovered almost all of the perplexity lost to one-shot pruning but only a quarter of the lost GSM8K accuracy, and our element-wise student matches Sparse-Llama's perplexity while solving half as many GSM8K problems. Sparse-Llama shows that reasoning does return after a much longer run on different data, but perplexity gives no sign of how far a run still has to go. Since retrained sparse models are commonly judged by perplexity, a reasoning task belongs in any judgment of whether retraining is done.

The floors therefore differ: on FP8, retraining at the scale of Sparse-Llama makes the speedup usable end to end; on Blackwell's sparse FP4, our 1B-token retraining leaves GSM8K far below the dense floor.
 Partial stacking does not avoid retraining either: sparsifying the units with the lowest predicted cost first beats random and kurtosis-based orders, but at half of the linear-layer FLOPs buys only 1.14--1.25\x{} (Appendix~\ref{app:stacking}).

%% file: tables/e2e.tex
\begin{tabular}{lccccc}
\toprule
 & weights & prefill & \multicolumn{2}{c}{decode (tok/s)} & \\
\cmidrule(lr){4-5}
serving & (GiB) & (k tok/s) & $b{=}1$ & $b{=}64$ & GSM8K \\
\midrule
dense BF16 & 15.0 & 41.4 & 136 & 6{,}424 & 0.596 \\
2:4 BF16 & 9.3 & 40.7 & 115 & 6{,}477 & 0.624 \\
\midrule
dense FP8 & 8.5 & 58.9 & 198 & 8{,}556 & 0.568 \\
2:4 FP8 & 6.1 & 74.6 & 234 & 9{,}914 & 0.612 \\
\bottomrule
\end{tabular}

%% file: sections/discussion.tex
\section{Conclusion}
\label{sec:conclusion}

At equal bits and equal tensor-core tiers, dropping one precision step was more accurate than adding 2:4 at every storage budget, and at every tier some dense format was both faster and more accurate, because the energy 2:4 discards along the forced axis does not shrink with bits. At the FP8 floor, a retrained 2:4 model served by vLLM on an H100 gained 27\% in prefill throughput over the dense FP8 model, with GSM8K accuracy on par. Whether 2:4 pays off at the floor is decided by retraining more than by kernels: Blackwell's pairwise sparse FP4 is fast but not yet usable at our budget, and perplexity, which recovered within a quarter billion tokens, is a poor guide to when retraining is done.

%% file: sections/appendix.tex
\section{Bit Accounting}
\label{app:bits}

\begin{table}[H]
\centering
\caption{Bits per original element ($g$ = 128 original elements per FP16 scale unless stated). Element-wise 2:4 stores a 2-bit index per kept value (1 bit per element); pairwise 4:8 stores two 2-bit pair indices per eight elements (0.5 bit per element) and one E4M3 scale per 32 positions, i.e.\ per 16 kept values.}
\label{tab:bits}
{%
\begin{tabular}{lll}
\toprule
Configuration & Formula & Bits \\
\midrule
Dense INT$k$, symmetric & $k + 16/g$ & 8.125 / 5.125 / 4.125 / 3.125 \\
Dense INT$k$, asymmetric & $k + 32/g$ & INT4: 4.25 \\
FP8, per-row or per-channel scale & $8 + \approx 0$ & 8.0 \\
MXFP4 (E8M0 scale per 32) & $4 + 8/32$ & 4.25 \\
NVFP4 (E4M3 per 16, FP32 per tensor) & $4 + 8/16$ & 4.5 \\
2:4 + FP16 & $8 + 1$ & 9.0 \\
2:4 + INT$k$ & $k/2 + 1 + 16/g$ & 5.125 / 4.125 / 3.125 \\
Sparse NVFP4, pairwise 4:8 (sm\_120) & $4/2 + 4/8 + 8/32$ & 2.75 \\
\bottomrule
\end{tabular}}
\end{table}

\paragraph{Budgets.} At 9/8.1 bits, 2:4 + FP16 meets INT8 and FP8; at 5.1 bits, 2:4 + INT8 meets INT5 (symmetric and asymmetric) and NVFP4; at 4.1 bits, 2:4 + INT6 meets INT4 (symmetric and asymmetric) and MXFP4; at 3.1 bits, 2:4 + INT4 meets INT3 (symmetric and asymmetric). Where costs cannot match exactly, the dense side gets fewer bits (FP8 and INT8 at 8.0 and 8.125 against 9, NVFP4 at 4.5 against 5.125), except MXFP4, which costs an eighth of a bit more than 2:4 + INT6. An ideal encoding of the six possible 2:4 patterns would need $\log_2 6 / 4 \approx 0.65$ bits per element instead of the hardware's 1 bit. Without scales, adding 2:4 to $k$-bit data compresses it by only $2k/(k+2)$.

\section{Predicted Ratio at Compute Tiers}
\label{app:tiers}

At a compute tier, both operands of the GEMM contribute error. Treating their relative errors as additive in the output, the 2\x{} and 4\x{} tiers give
\begin{align}
\Rpred^2 &= \frac{\etf(A)}{e_8(A) + e_8(B)} && \text{(2\x{})}, \label{eq:rrule_2x}\\
\Rpred^2 &= \frac{\etf(A) + e_8^{\text{kept}}(A) + e_8(B)}{e_4(A) + e_4(B)} && \text{(4\x{})}, \label{eq:rrule_4x}
\end{align}
where $B$ is the dense operand and $e_8$ and $e_4$ are the errors of the 8-bit and 4-bit formats. In the 2\x{} tier, the pruning loss of one operand must beat the 8-bit rounding error of two, and dropping half of the elements costs far more than rounding them to 8 bits.

\section{Measurement Details and Tables}
\label{app:atlas}

\paragraph{Models.} Llama-3.1-8B-Instruct (revision 0e9e39f) and Qwen3-8B (revision b968826) for the comparison above the floor; Llama-3.1-8B base (revision d04e592) and Sparse-Llama, RedHatAI/Sparse-Llama-3.1-8B-2of4 (revision c28c229, the general model, not a task-tuned variant), at the floor.

\paragraph{Samples and positions.} Each layer is evaluated on 16 samples: 8 LongBench documents~\citep{bai2024longbench} (two each from 2WikiMQA, HotpotQA, Qasper and GovReport, truncated to 16{,}384 tokens) and 8 C4 validation documents. Errors are measured at causal-safe query positions, the last position of each 128-token block. Both sides calibrate on 128 C4 sequences of 2048 tokens, with statistics from the uncompressed model. Linear-layer weights and inputs pool the four projection sites (fused QKV, output, fused gate/up, down). With grouped-query attention, keys and values are measured per KV head, queries and probabilities per query head. The measurements comprise 221{,}408 rows, half of them with Hadamard rotation.

\paragraph{Aggregation.} The layer-median $R$ of a cell is the median over layers of the per-layer median of $\log R$ over samples and projection sites; its 95\% interval comes from a paired bootstrap over samples with 10{,}000 resamples. \Rbasic{} is the same ratio with the basic method on each side (magnitude pruning against round-to-nearest). The criterion baseline is a logistic regression on $\log(\text{excess kurtosis} + 3)$, fitted on one model and tested on the other.

\begin{figure}[H]
\centering
\includegraphics[width=0.5\textwidth]{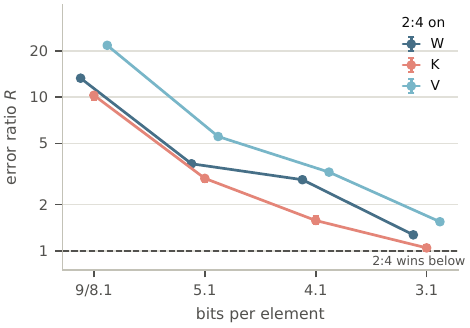}
\caption{Figure~\ref{fig:accuracy}a (equal bits) for Llama-3.1-8B-Instruct.}
\label{fig:bits_llama}
\end{figure}

\begin{table}[H]
\centering
\caption{Storage budgets: layer-median $R$ against the strongest quantizer (in parentheses).}
\label{tab:storage_all}
\input{tables/storage}
\end{table}

\begin{table}[H]
\centering
\caption{Compute tiers: layer-median $R$ per family and rival. Bold: $R < 1$; all six such cells face MXFP4 or row-wise INT8 for P. MXFP4 is simulated.}
\label{tab:compute_all}
\input{tables/compute}
\end{table}

\begin{table}[H]
\centering
\caption{How well the criterion separates the points where 2:4 wins at operator level ($\Rbasic < 1$). Always predicting quantization is as accurate as the criterion for W--V; the criterion's value lies in ranking (AUC, rank correlation), not in classification at the threshold. AUC and Spearman as in Figure~\ref{fig:criterion}; kurtosis fitted on the other model.}
\label{tab:criterion}
\resizebox{\textwidth}{!}{\input{tables/criterion}}
\end{table}

\section{Model-Level Checks}
\label{app:proxy}

A model-level cell compresses its family in every layer, each side using its most frequent winning method in the operator-level measurements; keys and values are compressed block-causally, with the current 128-token block kept dense. WikiText-2 perplexity uses 141 windows of 2048 tokens, each beginning with the BOS token, as every calibration sequence does (GPTQ calibrated on such sequences fits Llama's massive BOS activations~\citep{sun2024massive} and fails on windows without it); GSM8K uses the first 250 test problems with 8-shot chain of thought~\citep{wei2022cot}, greedy decoding and strict matching (95\% interval about $\pm$6 points); RULER uses five tasks at 16K tokens with 25 items each. Prompts are plain text without a chat template, so Qwen3's thinking mode is not engaged; GSM8K generates at most 256 tokens and stops at the next question, and RULER uses each task's own limit. Differences between the two sides are tested with a paired bootstrap over items (GSM8K, RULER) or windows (perplexity), 10{,}000 resamples; a difference that is not significant is not evidence of equivalence. On the fifteen boundary cells ($0.7 < R < 1.4$), operator-level $R$ agrees with perplexity on 4 of 7 Llama and 5 of 8 Qwen3 cells, with GSM8K on 4 of 7 and 6 of 7, and with RULER on 3 of 7 and 3 of 8. RULER was planned for keys, values and probabilities; the six weight, input and query cells were added later and run on L40S GPUs, both sides on the same machine. A configuration shared with an H100 run reproduces its RULER score within 0.6 points.

\begin{figure}[H]
\centering
\includegraphics[width=\textwidth]{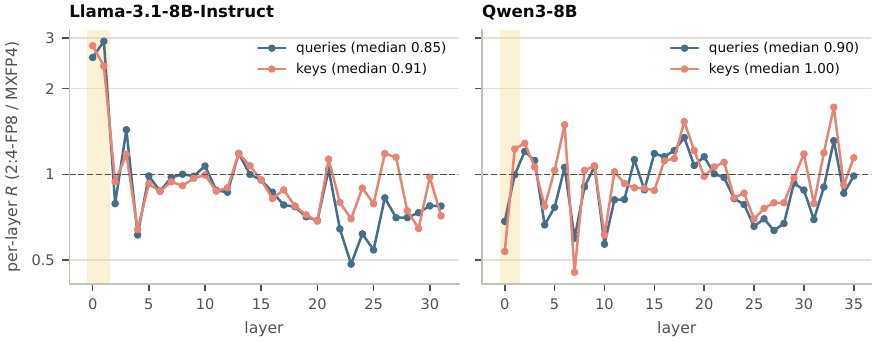}
\caption{Per-layer operator-level $R$ of queries and keys against MXFP4 (4\x{} tier), the cells whose median favours 2:4 while model quality does not. Shaded: the first two layers.}
\label{fig:layers}
\end{figure}

\begin{table}[H]
\centering
\caption{Operator-level $R$ against model-level perplexity (Llama-3.1-8B-Instruct, WikiText-2; uncompressed 7.23). Two cells per family, one among the most and one among the least favourable to 2:4. These cells were chosen and evaluated in the pilot (eight layers, L40S GPUs) before the full measurements, so their $R$ and perplexity differ slightly from Figure~\ref{fig:model}; for example, keys at 3.1 bits have $R$ = 0.97 over the pilot layers and 1.05 over all 32.}
\label{tab:proxy}
\input{tables/proxy}
\end{table}

\section{Hardware Measurements}
\label{app:hw}

We use PyTorch 2.10, CUDA 12.8 and cuSPARSELt 0.8.1. Times are means of 100 calls after 20 warm-ups. Linear shapes are the four Llama-3.1-8B projections at $M \in \{2048, 4096, 16384\}$ (12 shapes); Qwen3-8B shapes give ratios within 0.04. Floors: FP8 uses per-channel weight scales and dynamic per-token activation scales; NVFP4 uses E2M1 values, an E4M3 scale per 16 elements and an FP32 tensor scale; INT4 weights use GPTQ with groups of 128.

\begin{table}[H]
\centering
\scriptsize
\caption{Speed ratios at the same precision and at equal tier. W: sparse weights, kernel only. X: sparse activations, online preprocessing on both sides. 8\x{}: pairwise 4:8 NVFP4 against dense NVFP4.}
\label{tab:hw_tiers}
\resizebox{\textwidth}{!}{\input{tables/hw_tiers}}
\end{table}

\begin{table}[H]
\centering
\caption{Share of online preprocessing in each side's time (sparse activations).}
\label{tab:prepshare}
\input{tables/hw_prepshare}
\end{table}

\begin{table}[H]
\centering
\caption{Preprocessing kernels with L2 flushed: time in \textmu s and, in parentheses, compulsory bytes over time as a fraction of measured copy bandwidth. A copy reads and writes every byte, so a read-dominated kernel can exceed 1. HieraSparse's original TileLang kernel ran only on L40S.}
\label{tab:prep}
\resizebox{\textwidth}{!}{\input{tables/hw_prep}}
\end{table}

\begin{table}[H]
\centering
\caption{Attention shapes as standalone GEMMs per KV group (four query heads, head dimension 128), sequence length 4096 / 16384: speed ratio with kernel time only $\to$ with online preprocessing. Entries above 2 on L40S at length 4096 reflect slow dense kernels chosen by the library heuristic at these narrow shapes.}
\label{tab:attention}
\resizebox{\textwidth}{!}{\input{tables/hw_attention}}
\end{table}

\begin{table}[H]
\centering
\caption{Decode-shaped GEMMs ($M \in \{16, 128\}$), L2 flushed before every call. cuSPARSELt does not support 8-bit sparse activations with 16 rows.}
\label{tab:decode}
\input{tables/hw_decode}
\end{table}

\begin{figure}[H]
\centering
\includegraphics[width=\textwidth]{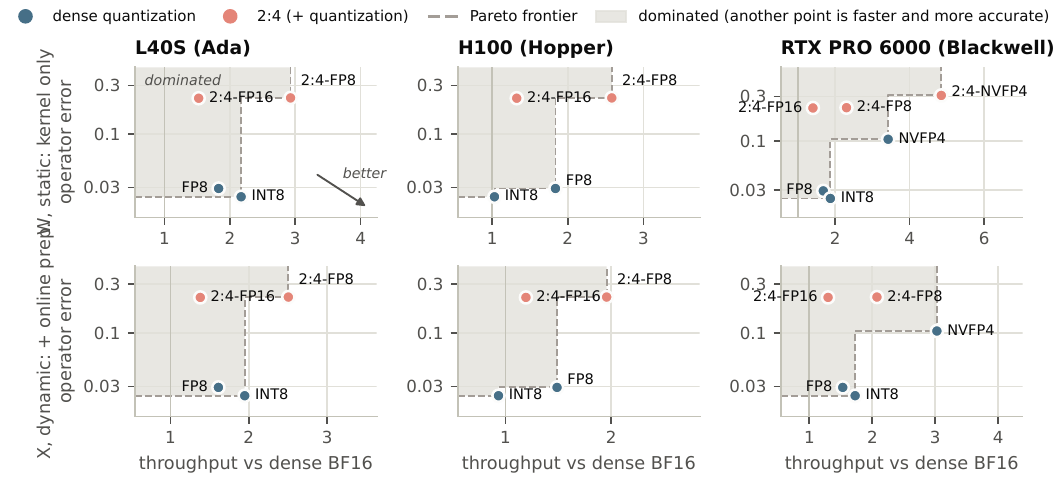}
\caption{Figure~\ref{fig:pareto} for Llama-3.1-8B-Instruct.}
\label{fig:pareto_llama}
\end{figure}

\begin{figure}[H]
\centering
\includegraphics[width=\textwidth]{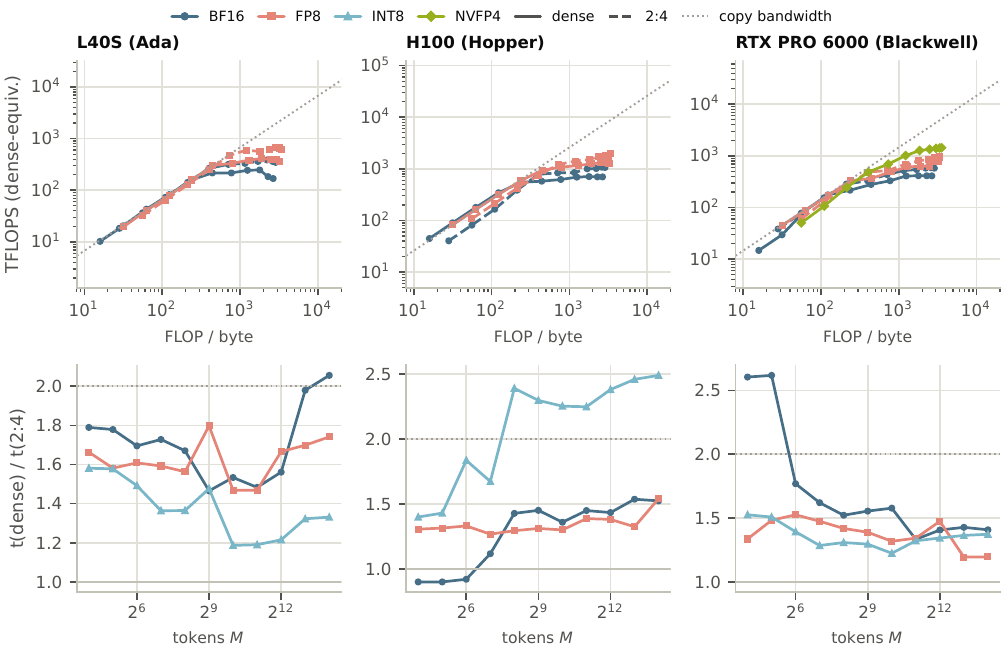}
\caption{Top: throughput against arithmetic intensity for the gate/up projection ($14336 \times 4096$, $M$ from 16 to 16{,}384), L2 flushed; solid dense, dashed 2:4; dotted, measured copy bandwidth. Bottom: 2:4 speedup over the same dense format against $M$.}
\label{fig:roofline}
\end{figure}

\section{Sparse FP4: Mask Pattern and Scale Granularity}
\label{app:fp4}

\begin{table}[H]
\centering
\caption{One-shot SparseGPT 2:4 on the weights at the NVFP4 floor (Llama-3.1-8B base, WikiText-2 perplexity), with the mask pattern and the scale block of the dense operand varied independently. Weights keep one E4M3 scale per 32 positions (16 kept values) throughout. Only the pairwise mask with 32-position scales runs on sm\_120; the other corners are simulated.}
\label{tab:fp4ablation}
\input{tables/fp4_ablation}
\end{table}

\section{Partial Stacking}
\label{app:stacking}

Units are (layer, projection) pairs. Each order makes the first units one-shot 2:4 until a share $f$ of the linear-layer FLOPs is sparse: by the cost the criterion predicts, by global kurtosis, or at random (five seeds).

\begin{figure}[H]
\centering
\includegraphics[width=\textwidth]{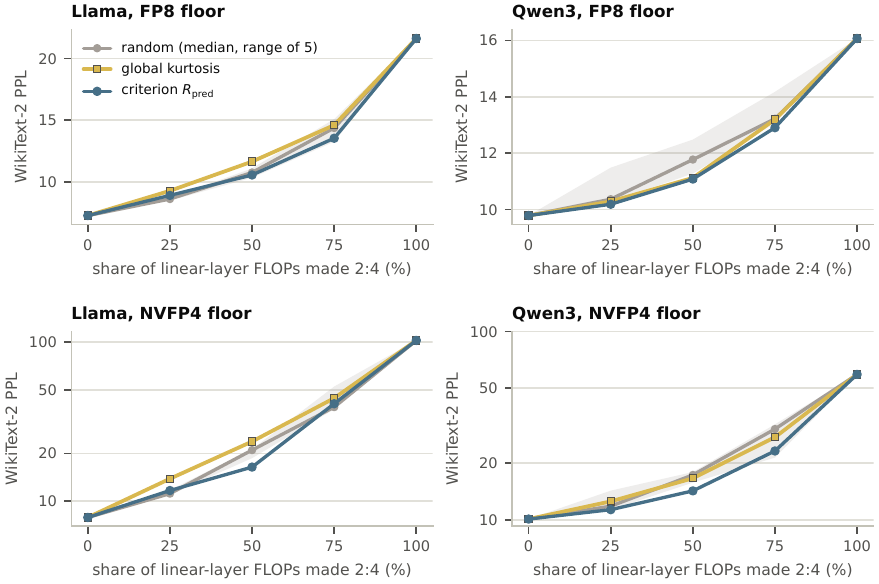}
\caption{Stacking one-shot 2:4 on the FP8 floor in the order given by the criterion, by global kurtosis, or at random.}
\label{fig:stacking}
\end{figure}

\section{Timing Details}
\label{app:timing}

\emph{Guarded profiling.} After other kernels have run in a process, a \texttt{torch.profiler} session can silently drop its first kernel events; short timings then come out too fast, and an algorithm search keeps configurations that appear to take no time. Every measurement therefore starts with 512 tiny pad kernels that absorb the lost events, and it is accepted only if each timed kernel is captured a whole number of times per call.

\emph{Clocks and libraries.} We record SM clock, power and throttle reasons before and after every run and use only GPUs that reach their normal clocks under load. With cuSPARSELt 0.8.1 preloaded, PyTorch's bundled 0.7.1 is also mapped into the process, so our \texttt{ctypes} calls bind the preloaded library explicitly and record its version.

\emph{Cache state.} An L2 cache of 50--128 MB holds whole activations at moderate $M$ and compressed weights at decode, so repeated calls run faster than a pipeline that reads from DRAM. Preprocessing, decode-shaped GEMMs and the roofline sweep are therefore also timed with L2 flushed before every call; the prefill conclusions hold under both.

\emph{End-to-end serving.} vLLM 0.11.2 is the last release that installs on our H100 host (glibc 2.28) and still serves 2:4 checkpoints, whose support was removed in 0.19~\citep{vllm36799}. LLM Compressor 0.8.1 writes the checkpoints in the compressed-tensors format: FP8 with per-channel weight scales and dynamic per-token activation scales on every linear layer except the LM head, and, for Sparse-Llama, the 2:4 bitmask format, which vLLM serves with its CUTLASS sparse GEMMs. Each configuration runs alone on one H100 with CUDA graphs, a 4096-token context, 85\% memory utilization and prefix caching off. Prefill and decode prompts are random tokens; each timing is the median of three runs after a warm-up, and decode ignores the end-of-sequence token. Weight memory is the model memory vLLM reports after loading and includes the BF16 embeddings and LM head, which 2:4 does not shrink. As a control, Sparse-Llama's weights were also served from dense-format checkpoints: throughput matches the dense rows of Table~\ref{tab:e2e} within 3\%, and GSM8K matches the 2:4 rows.

\section{Retraining Details}
\label{app:retrain}

\begin{table}[H]
\centering
\small
\caption{Retraining recipe.}
\label{tab:recipe}
\begin{tabular}{lp{13.5cm}}
\toprule
Start & One-shot SparseGPT on Llama-3.1-8B, all linear weights; pairwise 4:8 and element-wise 2:4 \\
Objective & Fixed-mask SquareHead distillation: KL divergence to the dense teacher on the next token plus the mean over layers of $\|h_s - h_t\|^2 / \|h_t\|^2$; pruned positions reset to zero after every step \\
Data & FineWeb-Edu sample-10BT, packed into 2048-token sequences with BOS per document; both students see the same sequences in the same order \\
Budget & 7{,}629 steps (999{,}948{,}288 tokens); checkpoints at steps 1{,}908, 3{,}815 and 7{,}629 \\
Optimizer & AdamW ($\beta$ = 0.9, 0.95; no weight decay), peak learning rate $2\times10^{-5}$, 100 warm-up steps, cosine decay to 10\%, gradient clipping 1.0; batch $64 \times 2048$ tokens; FSDP2 with FP32 master weights and BF16 compute \\
Hardware & 4 H100 per student, about 16 hours \\
\bottomrule
\end{tabular}
\end{table}

No hyperparameter was tuned, and each student was trained once. Format cost at 1B tokens (pairwise minus element-wise, BF16): +1.22 perplexity, 95\% interval [1.15, 1.29]; $-$0.14 GSM8K, interval [$-$0.20, $-$0.09].

%% file: tables/storage.tex
\begin{tabular}{llrrrr}
\toprule
Model & Family & 9 / 8.1 b & 5.1 b & 4.1 b & 3.1 b \\
\midrule
Llama & W & 13.27 (FP8-row) & 3.69 (NVFP4-g16) & 2.90 (INT4-g128) & 1.27 (INT3-g128) \\
Llama & K & 10.28 (FP8-row) & 2.96 (NVFP4-g16) & 1.58 (MXFP4-g32) & 1.05 (INT3-g128) \\
Llama & V & 21.71 (FP8-row) & 5.54 (NVFP4-g16) & 3.25 (INT4-g128) & 1.55 (INT3-g128-asym) \\
Qwen3 & W & 13.00 (FP8-row) & 3.61 (NVFP4-g16) & 2.77 (INT4-g128) & 1.22 (INT3-g128) \\
Qwen3 & K & 10.43 (FP8-row) & 2.62 (NVFP4-g16) & 1.41 (MXFP4-g32) & 1.27 (INT3-g128) \\
Qwen3 & V & 19.36 (FP8-row) & 5.13 (NVFP4-g16) & 3.10 (INT4-g128) & 1.49 (INT3-g128) \\
\bottomrule
\end{tabular}

%% file: tables/compute.tex
\begin{tabular}{llrrrr}
\toprule
Model & Family & 2$\times$ vs FP8 & 2$\times$ vs INT8 & 4$\times$ vs NVFP4 & 4$\times$ vs MXFP4 \\
\midrule
Llama & W & 7.50 & 7.70 & 2.13 & 1.48 \\
Llama & X & 7.70 & 9.25 & 2.21 & 1.48 \\
Llama & Q & 5.17 & 19.05 & 1.72 & \textbf{0.85} \\
Llama & K & 5.33 & 20.33 & 1.88 & \textbf{0.91} \\
Llama & V & 15.83 & 2.31 & 3.18 & 2.07 \\
Llama & P & 6.12 & \textbf{0.78} & 1.28 & \textbf{0.78} \\
Qwen3 & W & 6.84 & 5.00 & 1.86 & 1.33 \\
Qwen3 & X & 7.12 & 4.96 & 1.96 & 1.25 \\
Qwen3 & Q & 6.31 & 14.91 & 1.78 & \textbf{0.90} \\
Qwen3 & K & 6.52 & 16.57 & 1.92 & 1.00 \\
Qwen3 & V & 15.64 & 3.52 & 3.20 & 2.00 \\
Qwen3 & P & 5.04 & 1.13 & 1.02 & \textbf{0.64} \\
\bottomrule
\end{tabular}

%% file: tables/criterion.tex
\begin{tabular}{lrrrrrrrrrr}
\toprule
 & & 2:4 & \multicolumn{2}{c}{accuracy} & balanced & recall of & \multicolumn{2}{c}{AUC} & \multicolumn{2}{c}{$|$Spearman$|$} \\
\cmidrule(lr){4-5}\cmidrule(lr){8-9}\cmidrule(lr){10-11}
Points & $n$ & wins & always quant. & criterion & accuracy & 2:4 wins & criterion & kurtosis & criterion & kurtosis \\
\midrule
W, X, Q, K, V & 3400 & 3.2\% & 96.8\% & 96.8\% & 0.56 & 12\% & 0.945 & 0.776 & 0.93 & 0.10 \\
P & 272 & 47.4\% & 52.6\% & 66.2\% & 0.66 & 71\% & 0.773 & 0.627 & 0.54 & 0.29 \\
W--V, within 25\% of parity & 152 & 41.4\% & 58.6\% & 56.6\% & 0.49 & 5\% & -- & -- & -- & -- \\
\bottomrule
\end{tabular}

%% file: tables/proxy.tex
\begin{tabular}{llrrrc}
\toprule
Family & Cell & $R$ (operator) & PPL, 2:4 side & PPL, quantized side & Agree \\
\midrule
K & storage 3.1 b, vs INT3-g128 & 0.969 & 7.86 & 8.03 & yes \\
K & storage 9/8.1 b, vs INT8-g128 & 40.6 & 7.82 & 7.23 & yes \\
P & compute 4$\times$, 2:4-FP8 vs MXFP4 & 0.848 & 7.29 & 7.52 & yes \\
P & compute 2$\times$, 2:4-FP16 vs FP8 & 6.18 & 7.29 & 7.23 & yes \\
Q & compute 4$\times$, 2:4-FP8 vs MXFP4 & 0.777 & 10.80 & 9.34 & \textbf{no} \\
Q & compute 2$\times$, 2:4-FP16 vs INT8 & 17.4 & 10.65 & 7.23 & yes \\
V & storage 3.1 b, vs INT3-g128-asym & 1.58 & 9.17 & 7.30 & yes \\
V & storage 9/8.1 b, vs INT8-g128 & 40.8 & 9.16 & 7.23 & yes \\
W & storage 3.1 b, vs INT3-g128 & 1.27 & 22.98 & 12.37 & yes \\
W & storage 9/8.1 b, vs INT8-g128 & 52.6 & 21.36 & 7.23 & yes \\
X & compute 4$\times$, 2:4-FP8 vs MXFP4 & 1.50 & 14.44 & 9.69 & yes \\
X & compute 2$\times$, 2:4-FP16 vs INT8 & 9.07 & 14.14 & 7.34 & yes \\
\bottomrule
\end{tabular}

%% file: tables/hw_tiers.tex
\begin{tabular}{llccccccc}
\toprule
 & & \multicolumn{3}{c}{same precision} & \multicolumn{4}{c}{equal tier} \\
\cmidrule(lr){3-5}\cmidrule(lr){6-9}
GPU & operand & BF16 & FP8 & INT8 & 2$\times$/FP8 & 2$\times$/INT8 & 4$\times$/NVFP4 & 8$\times$/NVFP4 \\
\midrule
L40S & W, kernel & 1.63 {\scriptsize [1.49, 1.98]} & 1.65 {\scriptsize [1.52, 1.90]} & 1.27 {\scriptsize [1.14, 1.41]} & 0.89 {\scriptsize [0.78, 0.92]} & 0.72 {\scriptsize [0.62, 0.77]} & -- & -- \\
L40S & X, + prep. & 1.43 {\scriptsize [1.25, 1.95]} & 1.59 {\scriptsize [1.50, 1.75]} & 1.24 {\scriptsize [1.12, 1.35]} & 0.89 {\scriptsize [0.70, 0.92]} & 0.70 {\scriptsize [0.56, 0.78]} & -- & -- \\
H100 & W, kernel & 1.44 {\scriptsize [1.37, 1.57]} & 1.42 {\scriptsize [1.34, 1.57]} & 2.39 {\scriptsize [2.16, 2.62]} & 0.74 {\scriptsize [0.62, 0.78]} & 1.29 {\scriptsize [1.02, 1.39]} & -- & -- \\
H100 & X, + prep. & 1.32 {\scriptsize [1.18, 1.44]} & 1.31 {\scriptsize [1.24, 1.47]} & 2.03 {\scriptsize [1.84, 2.30]} & 0.80 {\scriptsize [0.68, 0.89]} & 1.28 {\scriptsize [1.18, 1.37]} & -- & -- \\
RTX PRO 6000 & W, kernel & 1.43 {\scriptsize [1.32, 1.75]} & 1.33 {\scriptsize [1.23, 1.50]} & 1.34 {\scriptsize [1.17, 1.42]} & 0.84 {\scriptsize [0.73, 0.92]} & 0.76 {\scriptsize [0.70, 0.85]} & 0.66 {\scriptsize [0.61, 0.70]} & 1.40 {\scriptsize [1.36, 1.46]} \\
RTX PRO 6000 & X, + prep. & 1.32 {\scriptsize [1.02, 1.67]} & 1.29 {\scriptsize [1.16, 1.57]} & 1.30 {\scriptsize [1.15, 1.40]} & 0.85 {\scriptsize [0.67, 0.90]} & 0.77 {\scriptsize [0.60, 0.82]} & 0.70 {\scriptsize [0.63, 0.75]} & -- \\
\bottomrule
\end{tabular}

%% file: tables/hw_prepshare.tex
\begin{tabular}{lcccc}
\toprule
GPU & 2:4-FP16 side & 2:4-FP8 side & dense FP8 side & dense NVFP4 side \\
\midrule
L40S & 0.04 {\scriptsize [0.01, 0.17]} & 0.11 {\scriptsize [0.03, 0.23]} & 0.06 {\scriptsize [0.02, 0.16]} & -- \\
H100 & 0.10 {\scriptsize [0.04, 0.14]} & 0.24 {\scriptsize [0.09, 0.34]} & 0.16 {\scriptsize [0.05, 0.26]} & -- \\
RTX PRO 6000 & 0.04 {\scriptsize [0.01, 0.13]} & 0.09 {\scriptsize [0.03, 0.18]} & 0.06 {\scriptsize [0.02, 0.17]} & 0.15 {\scriptsize [0.06, 0.23]} \\
\bottomrule
\end{tabular}

%% file: tables/hw_prep.tex
\begin{tabular}{llccccccc}
\toprule
GPU & operand & fused 2:4 & fused 2:4+FP8 & FP8 & NVFP4 & cuSPARSELt & PyTorch tile & HieraSparse \\
\midrule
L40S & P $65536\times16384$ & 6753 (0.74) & 3806 (1.10) & 4714 (1.01) & 4714 (0.86) & 26948 & 9777 & 5189 \\
L40S & V $128\times16384$ & 10 (1.00) & 12 (0.70) & 10 (0.91) & 15 (0.55) & 35 & 11 & 8 \\
L40S & X $4096\times4096$ & 48 (1.61) & 51 (1.28) & 48 (1.56) & 49 (1.29) & 396 & 56 & 54 \\
L40S & X $16384\times14336$ & 1269 (0.86) & 823 (1.11) & 978 (1.07) & 907 (0.98) & 5950 & 1961 & 1100 \\
H100 & P $65536\times16384$ & 1530 (0.84) & 1728 (0.63) & 1742 (0.71) & 2263 (0.47) & 47354 & 4796 & -- \\
H100 & V $128\times16384$ & 6 (0.45) & 10 (0.22) & 7 (0.33) & 14 (0.15) & 275 & 13 & -- \\
H100 & X $4096\times4096$ & 22 (0.93) & 32 (0.52) & 28 (0.69) & 41 (0.40) & 531 & 81 & -- \\
H100 & X $16384\times14336$ & 280 (1.01) & 401 (0.59) & 394 (0.69) & 500 (0.46) & 9369 & 1052 & -- \\
RTX PRO 6000 & P $65536\times16384$ & 2277 (1.01) & 1892 (1.02) & 2141 (1.03) & 1844 (1.02) & 15086 & 3330 & -- \\
RTX PRO 6000 & V $128\times16384$ & 5 (0.93) & 7 (0.54) & 6 (0.67) & 9 (0.40) & 26 & 9 & -- \\
RTX PRO 6000 & X $4096\times4096$ & 24 (1.47) & 29 (1.03) & 31 (1.10) & 25 (1.19) & 216 & 34 & -- \\
RTX PRO 6000 & X $16384\times14336$ & 461 (1.09) & 420 (1.01) & 454 (1.06) & 393 (1.05) & 3312 & 695 & -- \\
\bottomrule
\end{tabular}

%% file: tables/hw_attention.tex
\begin{tabular}{llccccc}
\toprule
GPU & family & same prec. FP16 & same prec. FP8 & 2$\times$ vs FP8 & 4$\times$ vs NVFP4 & prep.\ share \\
\midrule
L40S & Q & 0.83$\to$0.82 / 0.83$\to$0.82 & 0.96$\to$0.96 / 0.99$\to$0.99 & 0.84$\to$0.84 / 0.92$\to$0.92 & -- / -- & 0.01 / 0.00 \\
L40S & K & 0.84$\to$0.84 / 1.01$\to$1.01 & 1.04$\to$1.04 / 0.95$\to$0.95 & 1.03$\to$1.03 / 0.88$\to$0.88 & -- / -- & 0.01 / 0.00 \\
L40S & P & 3.90$\to$0.51 / 1.77$\to$0.38 & 1.55$\to$1.25 / 1.63$\to$1.31 & 0.72$\to$0.79 / 0.92$\to$0.74 & -- / -- & 0.87 / 0.78 \\
L40S & V & 1.02$\to$1.01 / 1.03$\to$1.03 & 2.94$\to$2.73 / 1.01$\to$1.01 & 0.41$\to$0.42 / 0.58$\to$0.59 & -- / -- & 0.01 / 0.00 \\
H100 & Q & 0.84$\to$0.80 / 0.88$\to$0.87 & 0.96$\to$0.95 / 0.98$\to$0.98 & 0.84$\to$0.87 / 0.82$\to$0.82 & -- / -- & 0.05 / 0.01 \\
H100 & K & 0.84$\to$0.82 / 0.86$\to$0.86 & 1.10$\to$1.09 / 1.14$\to$1.14 & 0.98$\to$0.98 / 0.95$\to$0.95 & -- / -- & 0.03 / 0.00 \\
H100 & P & 1.49$\to$0.44 / 1.73$\to$0.38 & 1.28$\to$0.93 / 1.50$\to$1.07 & 0.86$\to$1.04 / 0.85$\to$1.08 & -- / -- & 0.70 / 0.78 \\
H100 & V & 0.97$\to$0.93 / 1.01$\to$1.00 & 0.98$\to$0.96 / 1.02$\to$1.01 & 0.58$\to$0.62 / 0.53$\to$0.53 & -- / -- & 0.05 / 0.01 \\
RTX PRO 6000 & Q & 0.91$\to$0.89 / 1.15$\to$1.15 & 1.02$\to$1.01 / 1.06$\to$1.06 & 0.91$\to$0.93 / 1.03$\to$1.03 & 1.12$\to$1.11 / 1.00$\to$1.00 & 0.02 / 0.00 \\
RTX PRO 6000 & K & 0.72$\to$0.71 / 0.91$\to$0.91 & 1.01$\to$1.01 / 0.94$\to$0.94 & 0.75$\to$0.76 / 0.91$\to$0.91 & 1.04$\to$1.04 / 0.95$\to$0.95 & 0.01 / 0.00 \\
RTX PRO 6000 & P & 1.75$\to$0.40 / 1.76$\to$0.46 & 1.41$\to$1.18 / 1.60$\to$1.23 & 1.02$\to$1.01 / 0.91$\to$0.92 & 0.76$\to$0.93 / 0.90$\to$0.96 & 0.77 / 0.74 \\
RTX PRO 6000 & V & 1.23$\to$1.20 / 1.05$\to$1.05 & 1.30$\to$1.27 / 1.01$\to$1.01 & 0.72$\to$0.74 / 0.52$\to$0.52 & 0.71$\to$0.75 / 0.58$\to$0.59 & 0.03 / 0.00 \\
\bottomrule
\end{tabular}

%% file: tables/hw_decode.tex
\begin{tabular}{llccccc}
\toprule
GPU & operand & BF16 & FP8 & 2$\times$/FP8 & 4$\times$/NVFP4 & 8$\times$/NVFP4 \\
\midrule
L40S & W, kernel & 1.72 {\scriptsize [1.58, 1.79]} & 1.46 {\scriptsize [1.33, 1.70]} & 0.90 {\scriptsize [0.86, 1.04]} & -- & -- \\
L40S & X, + prep. & 0.98 {\scriptsize [0.90, 1.02]} & 1.03 {\scriptsize [1.02, 1.04]} & 0.62 {\scriptsize [0.61, 0.63]} & -- & -- \\
H100 & W, kernel & 0.77 {\scriptsize [0.63, 1.11]} & 1.15 {\scriptsize [0.95, 1.30]} & 0.76 {\scriptsize [0.67, 0.84]} & -- & -- \\
H100 & X, + prep. & 0.72 {\scriptsize [0.55, 0.92]} & 0.93 {\scriptsize [0.91, 0.95]} & 0.80 {\scriptsize [0.65, 0.84]} & -- & -- \\
RTX PRO 6000 & W, kernel & 1.63 {\scriptsize [1.20, 2.61]} & 1.28 {\scriptsize [0.98, 1.55]} & 0.85 {\scriptsize [0.71, 0.98]} & 1.35 {\scriptsize [0.99, 1.59]} & 1.65 {\scriptsize [1.21, 2.65]} \\
RTX PRO 6000 & X, + prep. & 1.02 {\scriptsize [0.97, 1.56]} & 0.95 {\scriptsize [0.83, 1.22]} & 0.67 {\scriptsize [0.58, 0.92]} & 0.92 {\scriptsize [0.71, 1.05]} & -- \\
\bottomrule
\end{tabular}

%% file: tables/fp4_ablation.tex
\begin{tabular}{lcc}
\toprule
Mask on the weights & activation scale per 16 & per 32 \\
\midrule
element-wise 2:4 & 21.74 & 22.39 \\
pairwise 4:8 & 105.23 & 108.83 (hardware) \\
\midrule
dense NVFP4 floor & 6.98 & -- \\
\bottomrule
\end{tabular}